# Machine Learning and Soil Humidity Sensing: Signal Strength Approach

LEA DUJIĆ RODIĆ, TOMISLAV ŽUPANOVIĆ, TONI PERKOVIĆ, and PETAR ŠOLIĆ (CORRESPONDING AUTHOR), University of Split, Croatia

JOEL J. P. C. RODRIGUES, Federal University of Piauí (UFPI), Teresina - PI, Brazil and Instituto de Telecomunicações, Portugal

The IoT vision of ubiquitous and pervasive computing gives rise to future smart irrigation systems comprising physical and digital world. Smart irrigation ecosystem combined with Machine Learning can provide solutions that successfully solve the soil humidity sensing task in order to ensure optimal water usage. Existing solutions are based on data received from the power hungry/expensive sensors that are transmitting the sensed data over the wireless channel. Over time, the systems become difficult to maintain, especially in remote areas due to the battery replacement issues with large number of devices. Therefore, a novel solution must provide an alternative, cost and energy effective device that has unique advantage over the existing solutions. This work explores a concept of a novel, low-power, LoRa-based, cost-effective system which achieves humidity sensing using Deep learning techniques that can be employed to sense soil humidity with the high accuracy simply by measuring signal strength of the given underground beacon device.

## 1 INTRODUCTION

Technologies of the 21st century, especially wireless technologies have been rapidly emerging in recent years. This has supported the development of Internet-connected sensory devices which provide observations and data measurements from the physical world. The expansion of large amount of connected devices has shaped a new paradigm- The Internet of Things (IoT). The Internet of Things represents a concept of ubiquitous computing technology such as sensors, actuators, mobile phones etc. that interact together prompted with the use of wireless technologies. In this environment it is possible to receive interesting and important information about or form the psychical world and moreover, interconnect it to exchange and use this information with the digital world [1]. This has had an impact on many areas of everyday life, particularly in fields such as automation, industrial manufacturing, logistics, business/process management, intelligent transportation of people and goods or environmental monitoring [2, 3]. IoT applications are especially suited for living environments like the agricultural ones regarding environmental sustainability or soil optimization where the irrigation

Authors' addresses: Lea Dujić Rodić, dujic@fesb.hr; Tomislav Županović, tzupan01@fesb.hr; Toni Perković, toperkov@fesb.hr; Petar Šolić (Corresponding Author), psolic@fesb.hr, University of Split, R. Boskovica 32, Split, Croatia, 21000; Joel J. P. C. Rodrigues, Federal University of Piauí (UFPI), Teresina - PI, Brazil, Instituto de Telecomunicações, Portugal, joeljr@ieee.org.

plays an extremely important role [4]. In agriculture, water issue and irrigation methods are essential in efficient water usage with regards to increase of productivity and economic benefit [5]. Determination of water status and conduction of irrigation could be resolved efficiently with the novel sensing technologies [6]. According to the study by California Department of Water Resources, and their evaluation of California weather- based "smart" irrigation controller programs, reduction of outdoor water may range from 6% up to 41% depending on the study site and region.[1] The crucial parameter in the development of a smart irrigation system is soil humidity which is affected by numerous environmental factors such as air temperature, air humidity and soil temperature [7].

There is a real need for the improvement of irrigation systems since it is estimated that 40% of water used for agriculture is lost in developing countries [8]. Wide variety of battery-operated sensing devices already exist which are based on measuring the electrical properties of the soil, while data is delivered through some wireless interface. Existing wireless technologies have been either designed for high-throughput applications (e.g., 3G, WiFi, LTE) with high power consumption or are characterized by low power consumption (e.g., ZigBee, Bluetooth Low Energy) but limited in the achievable coverage area. Low power wide area networks (LPWA), such as LoRa, Sigfox NB-IoT are emerging as the enabling wireless technology especially for the development of precision farming, flood monitoring [9], precision livestock farming and/or smart irrigation systems [10, 11]. LPWA leverage the need for only intermittent or sporadic transmissions of small data packets, making them suitable for battery-operated devices. Existing commercial sensors for irrigation systems are quite expensive, while sensor lifetime can reach up to couple of years.

The development of IoT has been followed by the exponential growth of big data [12] and with it arose Machine Learning (ML) with great potential for precise predictions made from the past observations given new measurements [13]. Within the field of smart agriculture, Machine Learning was proposed for moisture estimation and moisture prediction [14]. However, proposed Machine Learning based solution were not aimed at reducing the cost of sensing devices, but rather at predicting soil moisture using available sensing devices providing information about air humidity, temperature and soil temperature [7]. Deep Learning (DL) has recently been introduced in the field of agriculture as a modern and promising technique with growing popularity, since advancements and applications of DL offer much needed precision in this field, indicating its large potential [15]. Deep Learning models have the ability to provide better analytical insight of the IoT systems with two important advantages over traditional ML: reduced need for hand crafted data set and improved accuracy [16]. Recently, with advancement of Low power wide area networks such as LoRa, it was shown that there exists a correlation between RSSI and soil moisture, for sensors and gateways buried fully into the ground [17, 18]. In addition, RSSI signals from LoRa / LoRaWAN devices were already proposed for fingerprinting localization [19], as well as for secure key generation [20].

This paper explores a concept of cost-effective and low-power humidity sensing using Deep learning techniques. Namely, due to different radio propagation behaviour in wet conditions, the changes of soil moisture and the signal strength of buried device are tightly correlated [17, 18]. Approach presented in this paper reflects a more real-life scenario that considers soil humidity estimation system based on the signal strength data generated from an underground LoRa beacon, collected by multiple overground gateways, along with related Machine Learning techniques. Future humidity sensor-free device which implements a DL algorithm should result in a cost effective solution which is easier to maintain due to the prolonged battery lifetime.

---

[1] Evaluation of California weather-based "smart" irrigation controller programs https://p2infohouse.org/ref/53/52030.pdf, Accessed: 24 Februry, 2020.

This paper is structured as follows: In Section 3 implementation and architecture of LoRaWAN-based soil moisture sensor device is described. Section 4 gives preliminary analysis of collected data. Machine Learning approach to soil moisture estimation based on SVR and LSTM algorithm is given in Section 5. Finally, conclusion is given in Section 6.

## 2 STATE OF THE ART

IoT systems are becoming increasingly dynamic, complex and Machine Learning has been regarded as the key technology for prediction and estimation using models and algorithms [21]. These algorithms can detect patterns in large amounts of data, specifically in a time series of observational data [22]. Generally, Machine Learning methodologies include a learning process with a goal to extract knowledge from data (training data) in order to perform a task. The most prominent feature of a learning machine is that trainer of learning machine is ignorant of the processes within it.

Machine Learning has the ability to learn without being programmed for specific tasks [23] where learning algorithms can be categorized as Supervised, Unsupervised and Reinforcement learning.

- Supervised learning algorithms require external assistance by a supervisor and their aim is to learn the mapping from the input values x to the output values y where the correct values are provided by a supervisor [24].
- Unsupervised learning algorithms are only given input data and their goal is to find regularities between them [25].
- Reinforcement learning algorithms learn by making decisions based on actions that need to be taken in order to provide the most positive outcome [26].

Two most common ML tasks are *classification tasks* and *regression tasks*. They differ with regards to the type of desired outcome. In classification tasks the outcome is a discrete category as were in regression tasks the outcome is one or more continuous variables [1].

In recent years Deep Learning (DL) has been actively employed in IoT applications as one of Machine Learning approaches [16]. DL reliably mines real-world IoT data from noisy and complex environment in contrast to conventional ML techniques and is a strong analytical tool for huge data giving better performance for such tasks [27]. Traditional ML techniques depend on the quality and accuracy of the features given to the algorithm whereas DL can automatically represent and organize multiple levels of information to express complex relationships between data [28].

Application of machine learning techniques for agriculture use and specifically soil moisture estimation and prediction have interested researches for over two decades. One such research is presented in [29], where Soil moisture is estimated using remote sensing data from Tropical Rainfall Measuring Mission Precipitation Radar (TRMMPR). The aim of the study was the estimation of soil moisture content for the Lower Colorado River Basin area. The authors developed Support Vector Machine (SVM), Artificial Neural Networks (ANN) and multivariate linear regression (MLR) models for the estimation and showed that the SVM model is better in capturing interrelation between soil moisture, backscatter and vegetation in comparisons to the ANN and MLR models. Study presented in [30] investigated relationship between soil moisture, Precipitation-Evapotranspiration Index (SPEI) and climate indices in Xiangjiang River basin. Incorporating the climate impact on drought, the Support Vector Regression SVR model is built to predict the SPEI from climate indices. The results showed that the SVR model could improve prediction accuracy of drought in comparison to solely using the drought index as the only input parameter. In [31] authors used precipitation, daytime and nighttime land surface temperature, potential evapotranspiration (PET) estimated using mean temperature, the Normalized Difference Vegetation Index, the Normalized Difference Water Index data combined with large-scale climate indices and long-range forecast climatological data for drought forecasting in the area of South Korea. They developed Decision trees (DT), Random forest (RF) and Extremely randomized trees (ERT) models as both classification and regression models. Results show that the regression models gave better performance in majority of cases. Soil moisture estimation from meteorological data using SVR is given in [32]. The authors also compare the SVR model with ANN to validate capabilities of the SVR model and conclude that SVR outperforms ANN in all cases. A Deep Learning model for soil moisture estimation is presented in the study [33]. The authors use deep belief networks (DBN) to predict Soil moisture content form topographic properties, environmental and meteorological data such as evapotranspiration, leaf area index (LAI) and land surface temperature in the Zhangye oasis in Northwest China. A novel macroscopic cellular automata (MCA) model by combining DBN is given and compared with widely used neural network, multi-layer perceptron (MLP). The result show that the DBN-MCA model led to a reduction in root mean square error by 18% in comparison with the MLP model. The authors conclude that the MCA model is promising for modeling the temporal and spatial variations of Soil Moisture content.

Review of Machine Learning dedicated to applications of machine learning in agricultural production systems is presented in [22]. The authors emphasise key and unique features of popular ML models and conclude that the ML models will be even more widespread in the future providing production improvement. Generally, a variety of ML algorithms have been exploited for agricultural purposes, such as [34], where soil moisture, crop biomass and Leaf Area Index are estimated form X-band ground-based scatterometer measurements using two variants of Radial basis function neural networks (RBFNN) algorithms, namely conventional radial basis function neural network and generalized regression neural network (GRNN). Results show that good performance was obtained from both networks in retrieving soil moisture content. Furthermore, in [35] support vector regression (SVR) technique was used and compared with multi layer perceptron neural network (MLP NN) algorithm for soil moisture estimation using C-band scatterometer field measurements and considering various combinations of the input features (i.e., different active and/or passive microwave measurements acquired using various sensor frequencies, polarization, and acquisition geometries). The authors present a comparison of SVR model performance and the MLP NN model and conclude that the SVR provides higher accuracy in prediction for the given data sets and for all the input feature configurations. They imply that the SVR model has a better generalization ability than the MLP NN model, i.e., the SVR model is more capable to learn mapping that provides higher accuracy in the prediction of unknown real samples. Futhremore, in [36] Long Short-Term Memory (LSTM) has been applied to predict water depth in agricultural Hetao Irrigation District in arid northwestern China using monthly water diversion, evaporation, precipitation, temperature, and time as input data to predict water table depth. The model was evaluated using RMSE and coefficient of determination $R^2$. The authors conclude that the proposed model is suitable for predicting water table depth and especially can be used in areas with complex hydro-geological characteristics. In [37] neural networks, multiple regression, and fuzzy logic were used for spatial soil moisture retrieval using active microwave data. The study area was located in Oklahoma, USA and models sensitivity was estimated by measuring the change of RMSE when an input variable is added (or deleted) from the models. The obtained results suggest that soil texture and vegetation highly influence soil moisture retrieval. The authors conclude that the fuzzy logic and neural network models out-preformed multiple regression in terms of validation. Table 1 gives a short comparison of above mentioned researches regarding ML models for soil moisture and drought estimation and forecasting.

Table 1. Comparison table of Machine Learning models and applications for Soil moisture and drought estimation.

| Paper | Prediction model | Application | Best peformance model |
|---|---|---|---|
| A. Sajjad et al. [29] | SVM, ANN, MLR | Soil moisture estimation | SVM |
| Y. Tian et al. [30] | SVR, drought index | Prediction of agricultural drought | SVR |
| J. Rhee and J. Im [31] | DT, RF, ERT (classification and regression models) | Drought forecasting | regression |
| M. Gill et al. [32] | SVM, ANN | Soil moisture prediction | SVM |
| X-D.Song et al. [33] | DBN, MLP | Soil moisture content prediction | DBN |
| R. Prasad et al. [34] | RBFNN, GRNN | Soil moisture estimation | both |
| L. Pasolli [35] | SVR, MLP NN | Soil moisture estimation | SVR |
| J. Zhang et al. [36] | LSTM | Water depth | LSTM |
| T. Lakhankar et al. [37] | neural networks, multiple regression, fuzzy logic | Spatial soil moisture retrieval | fuzzy logic, neural network |

Soil moisture estimation from other measurements such as RSSI only recently started to attract the research community. In [38] authors present a case study of how variations in meteorological conditions with four selected meteorological factors, air temperature, absolute air humidity, precipitation and sunlight influence IEEE 802.15.4 network based on six months of sensor data. Amongst the obtained results, they conclude that temperature is the most dominantly correlated with RSSI. Similarly, in [39] an impact of both air temperature and air humidity on performances of signal strength variations of 802.15.4 networks is shown. The authors conclude that air temperature has a significant negative influence on signal strength in general, while high relative air humidity may effect the signal on lower temperatures.

In [40], authors present a soil moisture monitoring system that uses UHF RFID tags in order to provide a wireless and battery-less field sensor. The paper presents the conceptual design of the system and provides experimental results showing that RSSI signal correlates with soil moisture using ANN. Further on, Artificial Neural Network was used for soil moisture prediction based on the RFID tag signal analyses giving coefficient of determination $R^2$ > 0.9 in majority of cases. However, since buried UHF RFID tags can be only read from short distances (up to 50 cm), this work proposes a mobile robot that travels across the field and navigates above the buried UHF RFID tags to collect RSSI data. Such a solution is time consuming and challenging especially when a large number of RFID sensors is scattered over a large and possibly uneven crop field, which requires a robot to travel to every tag to collect RSSI data. In [41] authors propose a passive UHF RFID tags sensors integrated with a monopole probe for soil moisture monitoring. Their experimental results show that changes in soil permittivity cause changes in RSSI of the back-scattered signal. Therefore, we conclude that further exploration of Received Signal Strength in the context of soil moisture estimation in required, especially for the purpose of reducing size, cost and battery efficiency of sensor device. With regards to LoRa based systems, research community has recently began to research the potential of correlating RSSI changes in LoRa signal with specific environmental changes. In study presented in [19] a publicly available dataset of LoRaWAN RSSI measurements is utilized to compare different machine learning methods for fingerprinting localization. The authors present the k-Nearest Neighbours (kNN) method, the Extra Trees method and a neural network approach using a Multi-layer Perceptron. They conclude that the MLP performs best achieving highest accuracy. In [42], the authors present results of signal strenght measurements and simulations based on Wireless InSite radio propagation software and imply that parking occupancy can be estimated by detecting the change in RSSI at the receiver side. With regards to soil moisture and its correlation to RSSI researches, [17, 18] show existence of correlation between RSSI signals from LoRa-based devices and soil moisture, for sensors and gateways buried fully into the ground. In [18] authors present design and experimental validation of the developed Soil Moisture Sensing System (SoMoS) based on a Software Defined Radio (SDR) approach using LoRa in the laboratory. The system showed valid behavior and is able to detect soil moisture via the radio field. Furthermore, authors in [17] have done a long term evaluation of the previously proposed SoMoS system showing a high correlation between measured Receive Signal Strength Indicator and precipitation events.

Work presented in our paper reflects a more real-life scenario that considers soil humidity estimation based on the signal strength data generated from an underground LoRa beacon and collected by overground gateways, using related Machine Learning techniques. The long range nature of LoRa technology allows device to communicate over larger distances (as far as 10 km) in comparison to UHF RFID and 802.15.4 radio technology. Hence, a single overground gateway device could collect simultaneously signal strength measurements data from multiple underground beacons scattered over a large crop fields, and estimate soil humidity using related Machine Learning techniques. Table 2 compares the above mentioned papers and this paper in terms of used radio technology for variety of applications.

Table 2. Comparison table of various radio technologies and applications based on signal strength variations.

| Paper | Radio technology | Application | ML model | Best performance model | RSSI values mapped with specific soil moisture values |
|---|---|---|---|---|---|
| Aroca et al. [40] | UHF RFID (short range) | soil moisture prediction | ANN | ANN | YES (MSE=0.00152) |
| Hasan et al. [41] | UHF RFID (short range) | soil permitivity | none | none | NO |
| Anagnostopoulos et al. [19] | LoRa (long range) | localization | kNN, Extra Trees, MLP | MLP | / |
| Solic et al. [42] | LoRa (long range) | parking space occupancy | none | none | / |
| Wennerström et al. [38] | 802.15.4 (short range) | change of meterological factors | none | none | NO |
| Luomala et al. [39] | 802.15.4 (short range) | air temperature and air humidity | none | none | NO |
| Liedmann et al. [18] | LoRa (long range) | soil moisture estimation | none | none | NO |
| Liedmann et al. [17] | LoRa (long range) | soil moisture estimation | none | none | NO |
| This paper | LoRa (long range) | Soil moisture estimation | SVR, LSTM | LSTM | YES (MSE=0.00018) |

## 3 LORA-BASED SOIL MOISTURE SENSOR

Usually, soil moisture sensor devices are battery-operated sensor devices where humidity and temperature (soil moisture) sensor is connected to a microcontroller that periodically reads sensor values and sends them over the radio channel to the base station (or base stations), after which the sensor data is forwarded to the application for visualization (Figure 1). To preserve energy, battery-operated devices go to the sleep mode, extending the device lifetime (usually for a couple of years). Concept exploited in this paper is based on the idea of elimination of soil moisture sensor device, thereby reducing the overall cost of sensor devices. This would result in energy savings required for its powering. In this section a detailed implementation description of LoRa-based sensor device is given. First, implementation of LoRaWAN-based I2C soil-moisture sensor device is introduced, after which a novel concept of sensor device that achieves moisture sensing through signal strength is explored.

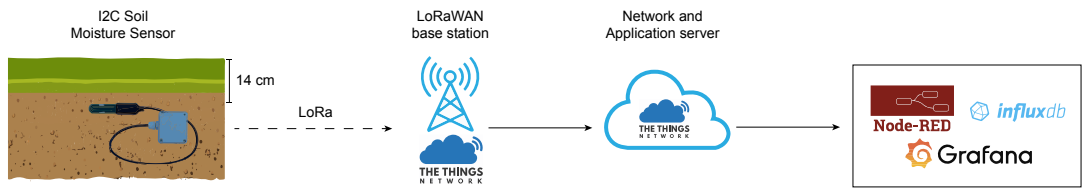

Fig. 1. Network architecture of LoRaWAN-based soil moisture sensor system.

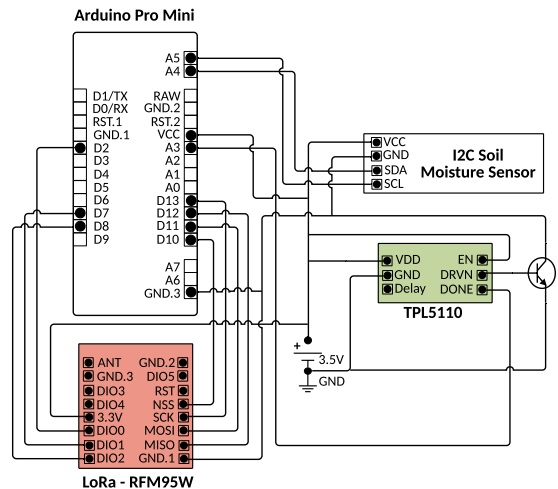

Fig. 2. Scheme of LoRaWAN I2C soil moisture sensor.

## 3.1 LoRaWAN Technology

In this paper the technology of Low Power Wide Area Networks (LPWAN) was employed, more precisely LoRa, to convey data over the radio from sensor device to the base station (Figure 1). Low-Power Wide Area Networks (LPWANs) such as LoRaWAN on allow battery-operated sensors or things to communicate low throughput data over long distances with minimal infrastructure deployment, and suited for applications in scenarios such as agriculture monitoring [17].

The network architecture of LoRaWAN typically exhibits a star-of-stars network topology, as depicted in Figure 1. The application specific end-devices are connected to one or many gateway or base station devices which are in turn directly connected to a network server and application. The gateways simply act as a transparent relay between the end devices and the network server. The network server can further forward the information to corresponding application servers for processing.

## 3.2 Realization of LoRaWAN-based I2C soil moisture sensor device

The core of sensor device is Arduino Pro Mini (with ATmega328P) that operates at 3.5V supply voltage, as depicted in Figure 2. For soil humidity and temperature monitoring an I2C soil moisture sensor is employed that uses capacitive sensing[2](with price up to 25 EUR), connected to Arduino Pro Mini board. To enable LoRaWAN-based communication, RFM95W module with SX1276 chip and spring antenna was used with +14dB transmission power. To preserve energy during inactive period, sensor device has a predefined sleeping period controlled with TPL5110 Nano Timer. During sleep period, TPL5110 basically cuts off power from both Arduino and sensors components, minimizing the overall consumption of the device. The timer was set to power up Arduino every 5 minutes.

Two Raspberry Pi gateway devices were used for implementation of LoRaWAN gateway infrastructure. These devices were placed indoor and configured to forward messages to The Things Network (TTN) cloud infrastructure. First gateway uses IMST iC880A-SPI concentrator with Shengda SDBF1.4 9dbi 868 MHz antenna, vertically polarized, while second gateway device uses RAK831 concentrator with Procom CXL 900-6LW-NB, 8 dBi gain, 868 MHz, vertically polarized, omnidirectional antenna.

---

[2]https://www.whiteboxes.ch/shop/i2c-soil-moisture-sensor/

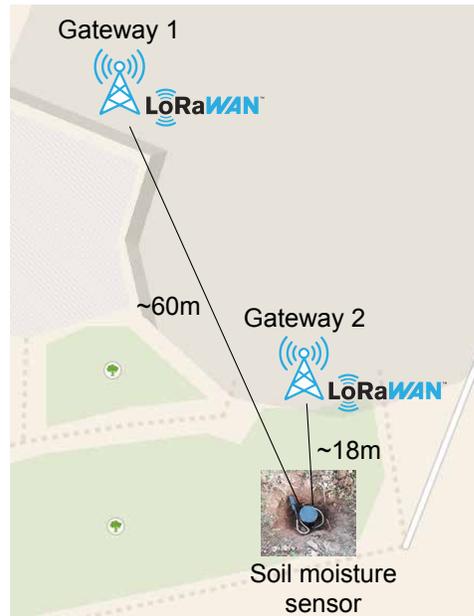

Fig. 3. Implementation of LoRaWAN-based soil moisture sensor.

The moisture sensor device was buried 14 cm below the ground level with antenna vertically oriented, while two indoor TTN gateways were placed in close vicinity of the sensor device. First gateway, denoted as Gateway 1 was placed on the fifth floor (15 meters from the ground and around 60 meters from the sensor), while the second gateway, denoted as Gateway 2 was placed at the first floor (around 4 meters from the ground and 18 meters from the sensor), as can be depicted in Figure 3. The sensor is buried to the specified depth of 14 cm for several reasons. The soil composition is brown soil on limestone and dolomite - calcambisol, which is developed on pure Mesozoic limestones and dolomites. The soil is non-carbonate with a pH greater than 5.5. The electrical conductivity of calcocambisol soil is about 124.2 $\mu$S/cm, with total porosity 45-65%, while the capacity of accessible water plants ranges from 5-15 cm [43]. As the aim of the work is to create a system / sensor for possible automatic irrigation, the depth at which the sensor is buried is large enough that the change in soil moisture affects the signal strength, while still being placed at depths from which roots of the plants draw water from.

Once the message arrives to the base station, it is forwarded to the TTN Network and Application server (Figure 1). Furthermore, TTN allows message forwarding from their infrastructure to our servers using MQTT protocol. On our server side, node-red was used for message aggregation, parsing and InfluxDB for storage. Figure 4 shows a snapshot of soil humidity and moisture along with RSSI and SNR values captured on one of the gateways. As can be seen, when the soil humidity increased, both RSSI and SNR signal values dropped, showing the tight bound between these values.

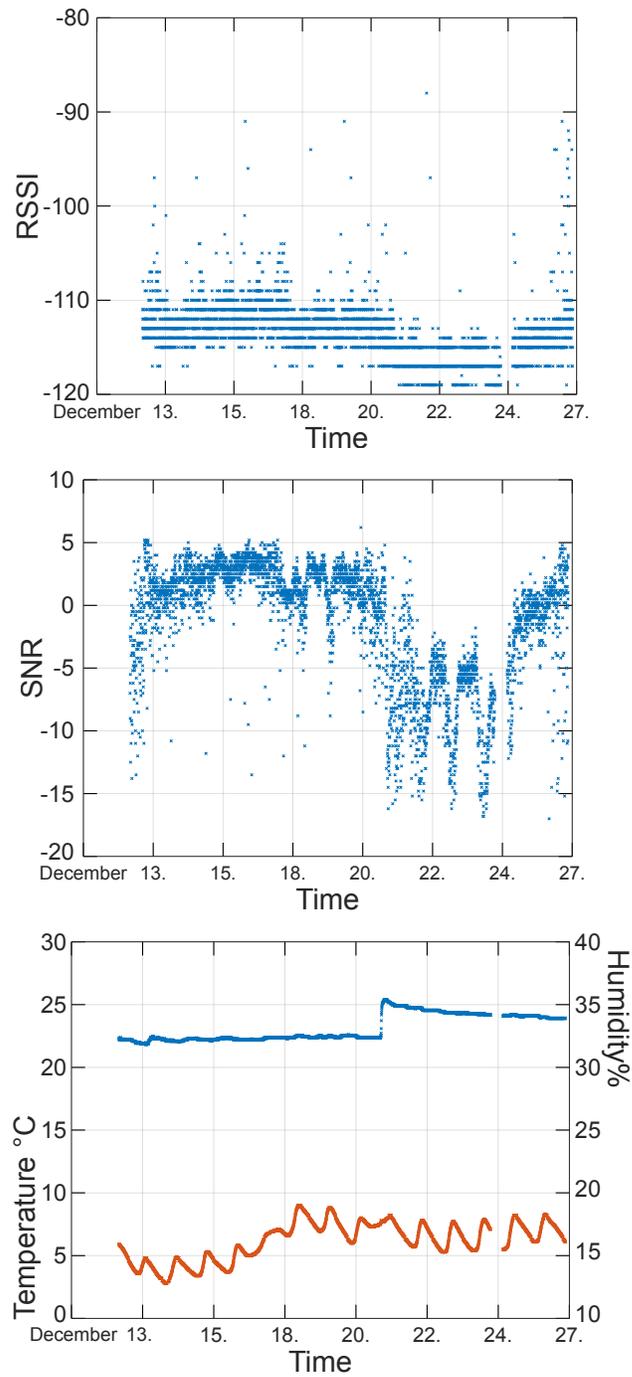

Fig. 4. Snapshot of RSSI and SNR signal captured on LoRaWAN gateways from soil moisture sensor. I2C sensor measures temperature and humidity of ground.

Table 3. Consumption of every element of the soil moisture sensor and beacon prototype along with lifetime duration estimation.

|  | Soil moisture device | Beacon device |
|---|---|---|
| LoRa RF96 IC | 116.1 mA | 116.1 mA |
| Soil moisture sensor v2.7.6 | 9.34 mA | 0 mA |
| Aduino mini pro (ATmega328p) | 9.09 mA | 4 mA |
| LDO | 0.00377 mA | 0.00377 mA |
| Timer TPL5110 | 0.000310 mA | 0.000310 mA |
| Active period duration | 7.5 s | 5.5 s |
| Average overall consumption in active period | 35 mA | 25 mA |
| Average consumption in inactive period | 4 $\mu A$ | 4 $\mu A$ |
| Lifetime duration (10.4 mAh battery) | 834.37 days | 1580 days |

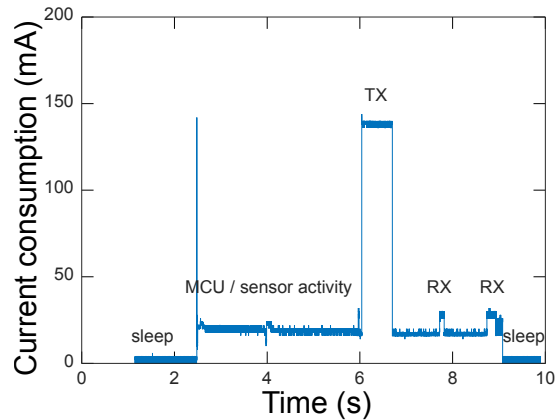

Fig. 5. Consumption of LoRaWAN I2C soil moisture sensor.

### 3.3 Consumption and lifetime estimation

Energy consumption of every component that comprises soil moisture sensor device is found in Table 3. To estimate the sensor lifetime the sensor was connected to the Current Ranger, whereas Current Ranger was connected to the Oscilloscope to capture detailed measurements of current consumption. As can be depicted in Figure 5, the biggest consumer of current is the LoRa module. After powering the device via TLP5110, Arduino MCU runs bootloader, reads LoRaWAN parameters from EEPROM memory. For I2C sensor heat and reading period around 1 second is spent, while bootloader additionally introduces 1 second for running. After that, message transmission occurs over LoRa radio channel. The transmission period was around 700 ms with 130 mA consumption. After that, two RX windows are open for message reception over the radio channel from the gateway (specified by LoRaWAN protocol). Next, MCU triggers TPL5110 to power off Arduino with radio and sensor components. While MCU is inactive, the overall consumption falls to 3.7 $\mu A$.

*Example:* Soil moisture sensors are equipped with built-in lithium-thionyl chloride (Li-SOCl2) batteries with an overall capacity of 10.4 Ah. Assuming that device consumption during the sleep period is 4 uA, while average device consumption during wake period is on average 35 mA, with 1 LoRaWAN keep-alive message sent every 10 minutes, and 7.5 seconds wake-up duration on average, an estimated battery lifetime will be approximately 834.37 days, or 2.29 years. This calculation assumes that the capacity is automatically derated by 15% from 10.4 Ah to account for the self discharge.

Table 4. Comparison of MCU consumption that support LoRaWAN communication.

| MCU | ATmega328P | ATtiny 84 | ATtiny 85 | STM32F103C8T6 (STM32) |
| --- | --- | --- | --- | --- |
| Consumption (3.3. V and 8MHz) | 3.9 mA | 3 mA | 3 mA | 8 mA |

**Consumption of Beacon Device.** In a concept of a low-power and cost-effective device where DL techniques could be employed to estimate soil humidity by measuring signal strength from underground beacon, a novel device will not require readings from I2C soil moisture sensor device. Hence, overall consumption will be reduced by saving around 1 seconds on I2C sensor heat and reading period. To further reduce consumption, additional 1 is saved by eliminating bootloader from Arduino. Therefore, the wake-up duration will now be reduced to approximately 5.5 seconds. To additionally decrease MCU consumption (ATmega328p) in active mode, all LEDs from Arduino Pro Mini could be taken off, lowering its consumption below 5 mA in active state. Table 3 depicts energy consumption of every component that builds Beacon Device. Taking into account the above presented numbers of battery capacity, wake-up duration and keep-alive messages every ten minutes, sensor lifetime will be increased to approximately 1580 days, or 4.33 years.

The consumption could be additionally reduced by introducing MCUs with lower consumption in active mode, besides ATmega328P. Table 4 gives comparison of consumption of MCU-s that support LoRaWAN library. Operating voltage 3.3V was selected since LoRa module operates at exact voltage, but also, smaller operating voltage reduces MCU consumption. Besides ATmega328P, MCUs that employ libraries for LoRaWAN-based connection are ATtiny 84, ATtiny 85 and STM32. Clearly, for the purposes of creating a simple beacon device, ATtiny 84 or ATtiny 85 could be used, as their consumption is around 3mA in active state.

In following section a soil moisture sensor concept is described that uses Deep learning technique based on signal strength measurement for soil humidity estimation.

## 4 PRELIMINARY ANALYSIS OF COLLECTED DATA FROM LORAWAN-BASED SOIL MOISTURE SENSOR

Analysis of data collected from LoRaWAN-based Soil Moisture Sensor device was conducted to uncover anomalies, define necessary data preparation approaches and determine potentially useful Machine Learning algorithms (ML algorithm) for the desired estimations. Such data analysis enabled the discovery of characteristic properties of the data with a goal to exploit how soil humidity is related to the signal strength.

### 4.1 Data Collection

The goal of the analyses was to detect how Soil humidity is related with Received Signal Strength Indicator. From the collected sensor data in InfluxDB, the data was extracted for the following time periods: 2 days of July, 17 days of August, 7 days of November, 31 days of December of 2019 and first 7 days of January 2020. Extracted information contained RSSI, SNR, soil temperature, soil humidity, a timestamp as well as the LoRaWAN Gateway ID. Namely, 13900 data packets were received by Gateway 1 (60 m distance), while Gateway 2 (18 m distance) collected 8413 data packets. Gateway 2 collected smaller amount of datapackets as it was not active all the time. In further text, RSSI and SNR from Gateway 1 are denoted as $RSSI_1$ and $SNR_1$, and from Gateway 2 as $RSSI_2$ and $SNR_2$.

### 4.2 Data Smoothing and Aggregation

Crucial features of data variables can be traced by observing their changes over time. However, as can be seen in Figure 4, Soil humidity changes slowly over time, while RSSI changes rapidly. Due to the channel stochastic behaviour the received signal has two major fading components. One is the rapid fluctuation of signal strength due to the propagation factors, whilst the other is its slower variant and a result of the multipath factors [44]. Therefore, raw data was smoothed through decomposition of the received signal strength on long and short term fading factors using a two-hour time frame. The long factor component was calculated by taking 24 samples of raw RSSI and SNR data and computing its mean, and further subtracted it from the raw value.

Data was then aggregated in such a way that the smoothed RSSI values were assigned to a class of Soil humidity percentage to determine if the RSSI values are correlated to a specific class of percentage of Soil humidity. The 13900 raw data of Soil humidity values vary form 10% up to 40% of humidity. However, the majority of data, i.e. 13570 (97%) are placed between values of 29% up to 39% of humidity. These values were categorized into equidistant classes which differ in 0.5% of humidity and RSSI values were associated to each class. Afterwards, for each class an RSSI mean value was assigned to it, where the mean was calculated out of all RSSI values in that class. Since 13900 data values of $RSSI_1$ were collected and 8413 data values of $RSSI_2$, analysis for each values were carried separately.

Pearson correlation coefficients between specific class of Soil Humidity and associated RSSI and SNR values is presented in Table 5.

Table 5. Pearson correlation matrix between soil humidity and RSSI and SNR.

|  | $RSSI_1$ | $SNR_1$ | $RSSI_2$ | $SNR_2$ |
|---|---|---|---|---|
| Soil humidity | -0.29 | -0.81 | -0.65 | -0.73 |

As can be seen in Table 5, $SNR_1$, $RSSI_2$ and $SNR_2$ substantially correlate with Soil humidity, implying that lower values of SNR and RSSI indicate higher soil humidity, as depicted in Figure 6.

It is important to note that for Gateway 1, that is more far away and on top of the University building (indoor), the channel was showing bi-modial distribution. Data tracing showed that there was particular working-day timeframe that changed mean value of the signal strength parameter. Difference between the results gained for $RSSI_1$ and $RSSI_2$ in data aggregation is a consequence of the distance of Gateway 1 (60 m) and Gateway 2 (18 m) form the soil moisture sensor. This implies that the farther the gateway is, the channel influences RSSI stronger than Soil Humidity. In light of this reasoning, it was concluded that RSSI and Soil Humidity values are considerably correlated and that the adequate Machine Learning algorithm must be able to comprise the complexity of the data properties elaborated in the above analyses.

## 5 MACHINE LEARNING APPROACHES TO SOIL MOISTURE ESTIMATION

The data analysis from previous section implied that the correlation between the Received Signal Strength Indicator. and soil humidity is substantial. Therefore, investigation if soil humidity can be estimated out of signal strength using Machine Learning methods is needed. Out of variety of different sets of data, the collected data is sequential data (i.e. time series) where all data points are not independent, and are identically distributed results from the time measurements. There are variety of ML models and methods for this kind of task such as ARIMA, Hidden Markov Model, Support

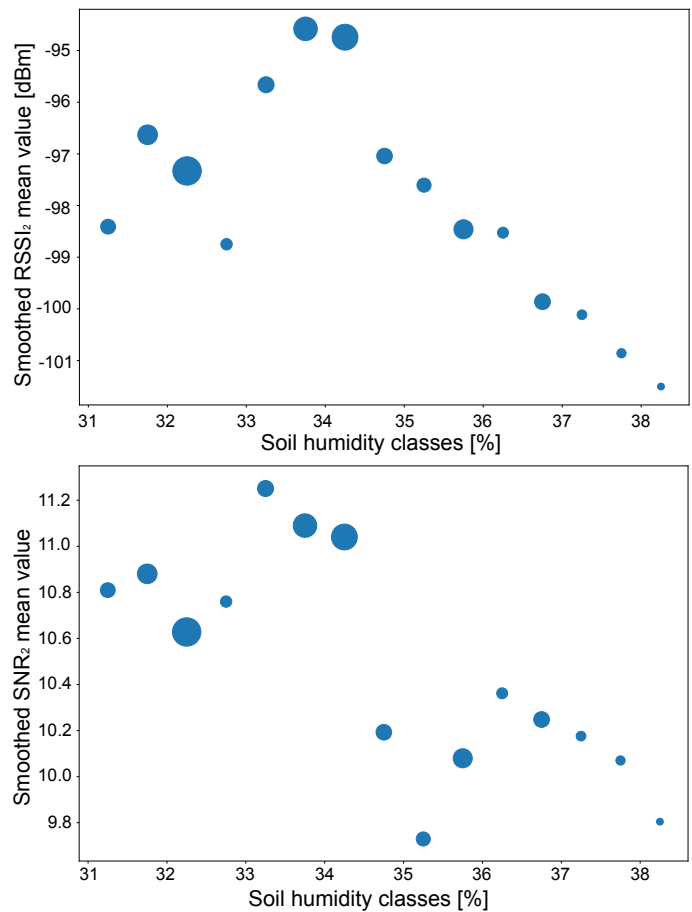

Fig. 6. Smoothed (up) $RSSI_2$ and (down) $SNR_2$ value correlated with specific Soil humidity classes. The circle size represents the relative portion of RSSI and SNR data within the humidity class.

Vector Regression (SVR), K-nearest-neighbours, or Recurrent Neural Networks (RNN). A strong advantage of Deep Learning for above mentioned problems is feature learning, i.e. the automatic feature extraction from raw data allowing it to solve more complex problems particularly well and fast [15].

Therefore, two models were built, evaluated and compared, namely: the SVR model and Long-Short Term Memory (LSTM) model. Both models used the same data and were validated in the same manner which is further described. The primary aim of building the models was not interpretation, but rather the accurate estimation of Relative Soil Humidity based on the signal strength. For the models, all 13900 raw data samples of RSSI and SNR captured on two LoRaWAN Gateways, as well as Soil Humidity were taken into account. Building the models consisted out of three steps: Data pre-processing, Defining the Model and Model validation, presented in the following.

*5.0.1 Data pre-processing.* Data pre-processing involved data normalization. This was done due the different value scales of variables in collected data, generally, Relative humidity was measured as a percentage, whereas RSSI and SNR values were measured in decibels. The inputs to the models were numerical values of RSSI and SNR, while the output was a numeric value that estimates relative Soil humidity. Furthermore, data was divided into training set and test set in 80 – 20% ratio respectively for models evaluation. The chosen training set consisted of pre-processed data from 2 days of July, 17 days of August, 7 days of November and 28 days of December of 2019 (containing 11120 samples); and test set consisted of pre-processed data of the last three days of December 2019 and first 7 days of January 2020 (containing 2780 samples). The part of data chosen for training was deliberately taken from above described time periods since it provided better insight into changes of RSSI with respect to Soil Humidity. Moreover, the test set was intentionally chosen from the previously described time frame because of consistency in measurements.

*5.0.2 Model validation.* The models were validated on the previously described test set. Loss function used for estimation of error was Mean Squared Error (MSE). Smaller MSE implies higher estimation accuracy, defined in equation:

$$MSE = \frac{1}{2m} \sum_{i=1}^{m} (\hat{y}^{(i)} - y^{(i)})^2. \tag{1}$$

As an additional evaluation metric that was used was Mean Absolute Error (MAE), defined in equation:

$$MAE = \frac{1}{m} \sum_{i=1}^{m} |\hat{y}^{(i)} - y^{(i)}|. \tag{2}$$

MSE gives average squared difference between the estimation and expected results whereas MAE measures the average magnitude of errors in a group of estimations. Moreover, validation loss represents how well or poorly the model behaves during training.

## 5.1 The SVR Model

The idea of Support Vector Machine (SVM) was introduced by Vapnik in mid 1990ties and today this a well known machine learning algorithm used in various applications from classification, forecasting to pattern recognition. The SVM implements the idea of mapping input vectors into a high-dimensional space $\mathcal{F}$, which is furnished with a dot product, using a non-linear mapping selected a priory [45]. This idea has been generalized to become applicable to regression problems using Support Vector Regression (SVR) briefly presented in the following.

Let us consider a training set $T = \{(x_i, y_i) \mid x_i \in \mathbb{R}^n, y_i \in \mathbb{R}, i = 1, ..., n\}$, where $X = (x_1, ..., x_n)$ are sampling data and $Y = (y_1, ...y_n)$ target vaules. The objective of SVR is to find function $f(x)$ that has at most $\varepsilon$- deviation from the observed target $y_i$ for all training data, enforcing flatness. This function can be defined as a linear function

$$f(x) = \omega \Phi(x) + b, \tag{3}$$

where $\Phi : \mathbb{R}^n \to \mathcal{F}$ is the map into the higher dimensional feature space, $\omega$ represents vector of wights of the linear function and $b$ is the bias. Desired function which is optimal is chosen by minimizing the function

$$\Psi(\omega, \xi) = C \cdot \sum_{i=1}^{n} \left(\xi_i + \xi_i^*\right) + \frac{1}{2} \|\omega\|^2, \tag{4}$$

where $\xi, \xi^*$ are non negative slack variables that measure the upper and lower excess deviation, $\|.\|$ is the Euclidean norm ($\frac{1}{2} \|\omega\|^2$ represents regularization term), $C$ is a regularization parameter which allows the tune of the trade-off between tolerance to empirical errors and regularization term. $\Psi(\omega, \xi)$ must satisfy following constraints:

$$\begin{cases} y_i - \omega\Phi(x_i) + b_i \leq \varepsilon + \xi_i \\ \omega\Phi(x_i) + b_i - y_i \leq \varepsilon + \xi_i^* \\ \xi, \xi^* \geq 0, i = 1, .., n \end{cases} \quad . \tag{5}$$

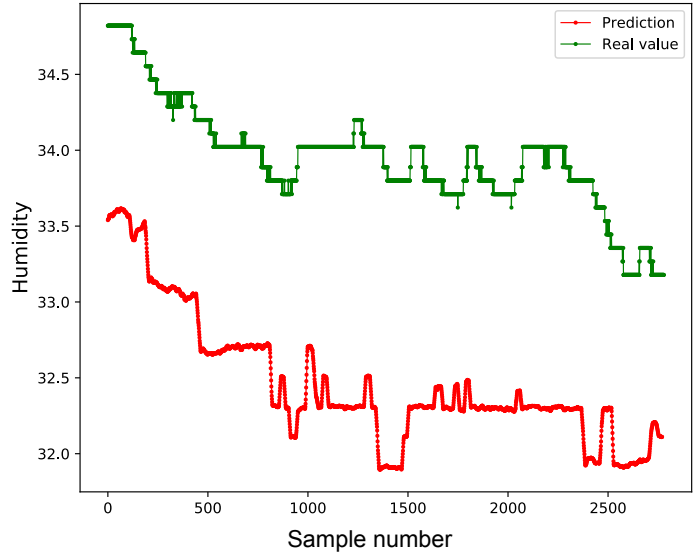

Fig. 7. Estimation of soil humidity with SVR model on the test set compared to expected values of soil humidity

Furthermore, the most prominent feature of SVR is the ability to establish correlation between data using non-linear mapping. This is achieved using kernel functions for generating the inner products, know as kernels, which satisfy Mercer's theorem. One of the broadly used kernels are polynomial and Gaussian radial basis function (RBF) kernels. Our model implemented the RBF kernel given with the formula: This is achieved using kernel functions for generating the inner products, know as kernels. Our model implemented the RBF kernel given with the formula:

$$K_\gamma(|x - x_i|) = \exp\left\{-\gamma \cdot |x - x_i|^2\right\} \tag{6}$$

Data was pre-processed as described, but the input for the model consisted out of the following: for every value of RSSI and SNR at the time step $t$, needed for the the estimation of humidity at a time step $t$, values of RSSI and SNR at the time step $t - 1$ were also taken. This provided the model with a "hybrid short-term memory" of the previously measured values in time $t - 1$. The necessary parameters $\gamma$, $C$ and $\varepsilon$ of the model were selected with Grid search process of performing hyper parameter tuning in order to determine the optimal values for a given model. The process resulted in parameters $\gamma = 1$, $C = 0.0.1$ and $\varepsilon = 0.1$.

Estimation of soil humidity with the model on the test set compared to expected values of soil humidity is presented in Figure 7.

The model was further validated on the test set resulting in $MSE = 0.0243$ and $MAE = 0.0487$ losses. This results implied that the Soil Humidity could be estimated based solely on Received Signal Strength and SNR values with a respectful accuracy even with a limited data set.

## 5.2 The LSTM Model

Recurrent Neural Networks are based on the recursive structure in which the one-step model with a time-step is trained first and then recursively used to return the multi-step prediction [46]. A special type of RNN is Long-Short Term Memory (LSTM) neural network constituted out of a set of recurrently connected memory blocks – LSTM cells (depicted in Figure 8). LSTM cell consists out of four layers, main layer and three layers which are gate controllers each computing values between 0 and 1 based on their input [16].

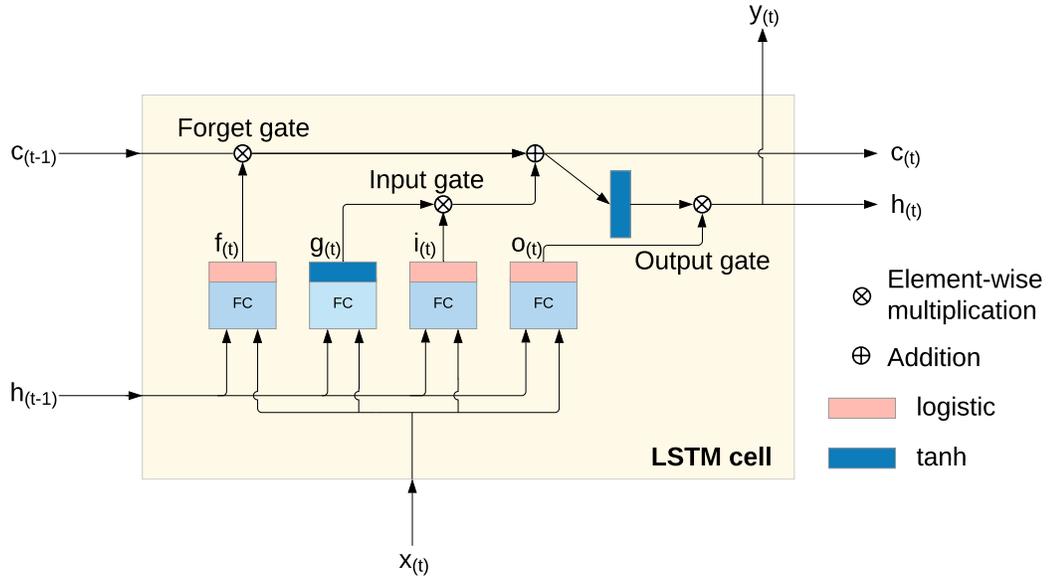

Fig. 8. Long Short-Term Memory (LSTM) cell.

Layers operate in the following way:

1. Main layer - analyses the current inputs $x_{(t)}$ and the previous (short-term) state $h_{(t-1)}$ then outputs the $g_{(t)}$ vector;
2. Forget gate $f_{(t)}$ decides parts of the long-term state $c(t-1)$ that need to be erased;
3. Input gate $i_{(t)}$ controls parts of $g_{(t)}$ that are added to the long-term state $c_{(t)}$;
4. Output gate $o_{(t)}$ determines which parts of long-term state should be read $c_{(t-1)}$ and given to the output y(t) and short-term state $h_{(t)}$ at the current time step(t).

The states of the cell are calculated using equations given below:

$$i_{(t)} = \sigma(W_{xi}^T \cdot x_{(t)} + W_{hi}^T \cdot h_{(t-1)} + b_i) \tag{7}$$

$$f_{(t)} = \sigma(W_{xf}^T \cdot x_{(t)} + W_{hf}^T \cdot h_{(t-1)} + b_f) \tag{8}$$

$$o_{(t)} = \sigma(W_{xo}^T \cdot x_{(t)} + W_{ho}^T \cdot h_{(t-1)} + b_o) \tag{9}$$

$$g_{(t)} = tanh(W_{xg}^T \cdot x_{(t)} + W_{hg}^T \cdot h_{(t-1)} + b_g) \tag{10}$$

$$c_{(t)} = f_{(t)} \otimes c_{(t-1)} + i_{(t)} \otimes g_{(t)} \tag{11}$$

$$y_{(t)} = h_{(t)} = o_{(t)} \otimes tanh(c_{(t)}) \tag{12}$$

where $\sigma$ represents logistic activation function, $tanh$ is hyperbolic tangent function, $W_{(x)}$ are weight matrices for each of the four layers for input vector $x_{(t)}$, and $W_{(h)}$ are matrices of the previous short-term state $h_{(t-1)}$. Finally, $b$ denotes the bias term of each layer. Difference between the LSTM and the standard RNN is within their structure to memorize. With traditional RNN parts of information are lost in the process of each feedback resulting in RNN not being able to have long time memory in contrast to LSTM which has a long term memory. LSTM is able to remove or add information to the cell state, unlike the mechanism that completely overrides cell states like in standard RNN [23]. Long dependency in time can be observed in IoT applications such as environmental monitoring, human activity recognition, or machine translation and LSTM models have proven to perform better than RNN for such data [16]. LSTM cells are very successful at capturing long-term patterns in time series data and that was one the reasons for their selection as Deep Learning approach for prediction.

The LSTM model presented in this paper used previously described pre-processed data. With regards to the inputs, a time step of 18 was chosen, approximating 90 minutes observations (18 samples 5 minutes period) for each estimation. Normalized data was given as an input to LSTM model with a goal to estimate relative Soil Humidity based on the signal strength. The architecture of model, presented in Figure 9, consisted of 2 stacked LSTM cells, on top of each other, which are able to build up progressively higher level representations of data.

Table 6. Selection of the hyper parameters for evaluation.

| Hyper parameter | Values |
| --- | --- |
| Number of neurons | 16, 32 |
| Learning rate | 0.0001 , 0.001 |
| Number of epochs | 50, 100, 150, 200, 250 |

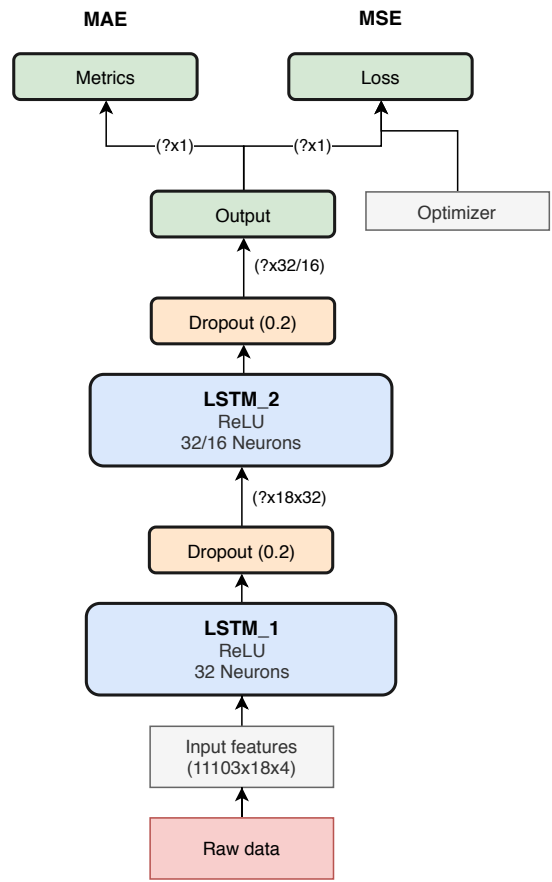

Fig. 9. Model architecture comprising two LSTM stacks.

Several options for number of neurons per layer, learning rates, epochs and different optimizers, presented in Table 6, were evaluated and compared. Table 7 shows results of different combinations of LSTM hyper parameters and their respective MSE and MAE losses. Although LSTM neural networks deal with time series data quite well, it is still rather important to cautiously regard selection of the hyper parameters. For example, reduced number of epochs decreases over fitting. To minimize the chance of over fitting, dropout layer was used after every LSTM cell layer with 20% probability of ignoring neurons throughout the training phase. Function used for activation was ReLU. With regards to selection of optimizers, a preliminary testing of model's behaviour was done with regards to different optimizers. Three optimizers were tested: Adam, RMSprop and SGD. RMSprop outperformed other two optimizers with regards to the aforementioned data set and was selected as optimizer.

Table 7. Results of different combinations of LSTM hyper parameters and their respective MSE and MAE losses

| Layer1 | Layer2 | Learning rate | Epochs | MSE | MAE |
| --- | --- | --- | --- | --- | --- |
| 32 | 16 | 0.0001 | 50  | 0.00332 | 0.05706 |
| 32 | 16 | 0.0001 | 100 | 0.01185 | 0.10531 |
| 32 | 16 | 0.0001 | 150 | 0.01266 | 0.11146 |
| 32 | 16 | 0.0001 | 200 | 0.00918 | 0.09515 |
| 32 | 16 | 0.0001 | 250 | 0.00320 | 0.05548 |
| 32 | 16 | 0.001  | 50  | 0.00126 | 0.03053 |
| 32 | 16 | 0.001  | 100 | 0.00255 | 0.04628 |
| 32 | 16 | 0.001  | 150 | 0.00032 | 0.01502 |
| 32 | 16 | 0.001  | 200 | 0.00119 | 0.03248 |
| 32 | 16 | 0.001  | 250 | 0.00089 | 0.02336 |
| 32 | 32 | 0.0001 | 50  | 0.00347 | 0.05664 |
| 32 | 32 | 0.0001 | 100 | 0.00215 | 0.03704 |
| 32 | 32 | 0.0001 | 150 | 0.00086 | 0.02409 |
| 32 | 32 | 0.0001 | 200 | 0.00562 | 0.07412 |
| 32 | 32 | 0.0001 | 250 | 0.00485 | 0.06734 |
| 32 | 32 | 0.001  | 50  | 0.00116 | 0.03149 |
| **32** | **32** | **0.001** | **100** | **0.00018** | **0.01043** |
| 32 | 32 | 0.001  | 150 | 0.00056 | 0.01460 |
| 32 | 32 | 0.001  | 200 | 0.00287 | 0.05083 |
| 32 | 32 | 0.001  | 250 | 0.00123 | 0.03421 |

LSTM model was evaluated on test set to assess the influence of different hyper parameters to the model performance. The best result was achieved with set of parameters with same number of neurons, 32, on both LSTM layers with a learning rate of 0.001 and number of epochs 100. As can be seen from the Table 7 they obtained the lowest MSE and MAE errors, 0.00018 and 0.01043, respectively. Figure 10 (a) represents the learning path of model with previously described parameters, training and validation loss in relation to each epoch. Estimation of soil humidity with the model compared to expected values of soil humidity that was done on the test set is presented in Figure 10 (b). Furthermore, it is important to emphasize the performance of the best LSTM model in terms of training and test time. The specification of the computational machine includes Intel core i5-7300HQ@2.50GHz processor, 8GB of RAM and NVIDIA GTX1050 GPU running 64 bit Windows 10 operating system and the NVIDIA CUDA Deep Neural Network library (cuDNN). The Keras 2.3.1. Python library was used running on top of a source build of Tensorflow 2.2.0 with CUDA support.me for different batch sizes. The best LSTM model training time was 1385.9917 seconds where as the test time was 0.66786 seconds, therefore causing minimum delay between two consecutive estimations. Since the computing was done on a desktop computer it is expected that the estimation time will be accelerated on a future dedicated computation machine using GPU. Due to their massively parallel architecture GPUs provide effective solutions for real-time systems, allowing them to speed up computations that involve matrix-based operations, which are the core of Machine learning implementations [47].

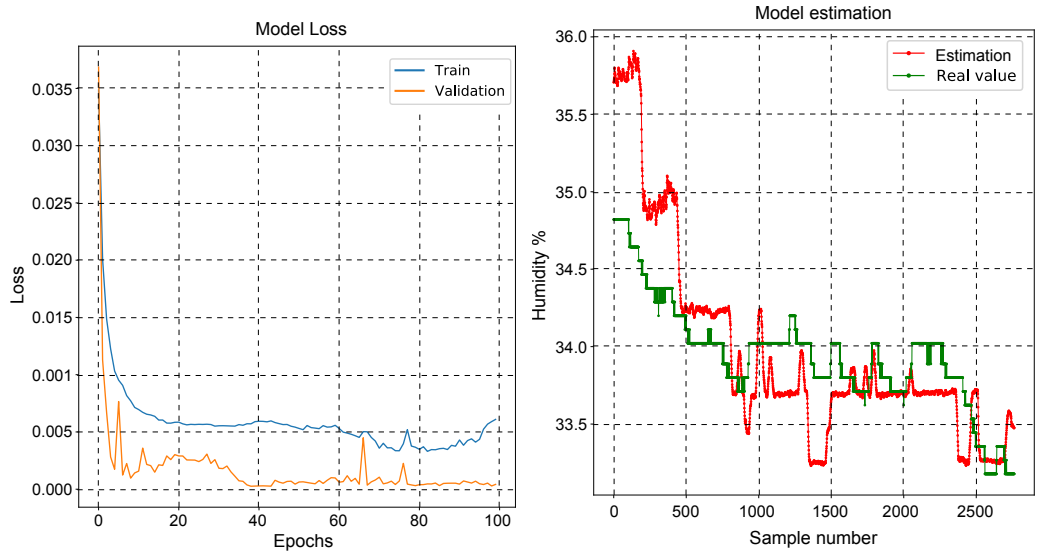

Fig. 10. (a) Learning path of model with training and validation loss, (b) estimation of soil humidity with the model compared to expected values of soil humidity.

*5.2.1 Discussion.* Despite a small and limited data set, containing only several months of representative data, significant results were gained with respect to the Machine learning approach for estimation of Soil Humidity based on the signal strength. Firstly, the SVR model gave considerably good estimation of Soil Humidity from RSSI and SNR indicating that the previously perceived correlation between SNR, RSSI and Soil Humidity was substantial. Secondly, the SVR model confirmed the Signal Strength Approach and concept of humidity sensing using only RSSI and SNR and applying Machine Learning techniques data was valid. Thirdly, the stacked LSTM model gained significantly more accurate estimations of soil humidity using this data and out preformed the traditional SVR with regards to the accuracy of estimation. Furthermore, LSTM, as a Deep Learning model that is suited for time series data, was able to better encompass complex correlation between the RSSI, SNR and Soil Humidity providing higher performance and precision.

## 6 CONCLUSION AND FUTURE DIRECTIONS

This paper presents a novel concept of cost-effective and low-power sensor that achieves humidity sensing using Deep Learning. Namely, LoRa-based I2C soil moisture sensor device was implemented that measures soil humidity and temperature for a time period of several months. With a sampling rate of 5 minutes, soil temperature and humidity were measured and sent over a radio channel to two LoRaWAN gateway devices that collected signal strength measurements from sensor device. The analysis of collected data showed noticeable correlation between RSSI, SNR and Soil Humidity. It was further shown that soil humidity can be estimated with high accuracy from signal strength, along with related Machine Learning techniques. Use of Long Short-Term Memory (LSTM) Neural Network as a Deep Learning approach provided significant results in terms of accuracy of estimation in contrast to traditional ML techniques of Support Vector Regression. Future work will comprise an improvement on the accuracy of presented LSTM model by involving other sensor measurements such as air humidity and temperature which could influence soil humidity estimation accuracy from signal strength. Moreover, potentials of energy savings will be examined. Transmission period of LoRa beacon using adequate machine learning algorithms will be investigated using dynamic wake-up and beacon transmissions that depends on dynamics of soil moisture changes.

## ACKNOWLEDGMENTS

This Technology Transfer Experiment has received funding from the European Union's Horizon 2020 research and innovation programme under the TETRAMAX grant agreement no 761349.

This work was partially supported by the Croatian Science Foundation under the project "Internet of Things: Research and Applications", UIP-2017-05-4206; by FCT/MCTES through national funds and when applicable co-funded EU funds under the Project UIDB/EEA/50008/2020; and by Brazilian National Council for Research and Development (CNPq) via Grant No. 309335/2017-5.